\documentclass[journal]{IEEEtran}

\usepackage{algorithm}
\usepackage[english]{babel}
\usepackage{blindtext}
\usepackage{algorithmic}

\usepackage{multirow}
\usepackage{xcolor}
\usepackage{bm}

\usepackage{amsfonts}
\usepackage{overpic}
\usepackage{graphics}
\usepackage{epsfig}
\usepackage{graphicx}
\usepackage{amsmath}
\usepackage{amssymb,subfigure,cases}
\usepackage{color,hyperref}

\ifCLASSINFOpdf
\else
\fi
\definecolor{darkblue}{rgb}{0.0,0.0,1.0}
\hypersetup{colorlinks,breaklinks,
            linkcolor=darkblue,urlcolor=darkblue,
            anchorcolor=darkblue,citecolor=darkblue}

\begin{document}

\title{Patch Alignment Manifold Matting}

\author{Xuelong~Li,~\IEEEmembership{Fellow,~IEEE,}
        Kang~Liu,~\IEEEmembership{Member,~IEEE,}
         Yongsheng~Dong,~\IEEEmembership{Member,~IEEE,}
        and~Dacheng~Tao,~\IEEEmembership{Fellow,~IEEE}


\thanks{This work was supported
in part by the National Natural Science Foundation of China under Grant
61761130079 and Grant U1604153, in part by the Key Research Program
of Frontier Sciences, CAS under Grant QYZDY-SSW-JSC044, in part by
the International Science and Technology Cooperation Project of Henan
Province under Grant 162102410021, in part by the State Key Laboratory
of Virtual Reality Technology and Systems under Grant BUAA-VR-16KF-
04, in part by the Key Laboratory of Optoelectronic Devices and Systems
of Ministry of Education and Guangdong Province under Grant GD201605,
and in part by the Australian Research Council under Project FT-130101457,
Project DP-140102164, and Project LP-150100671. (Corresponding author:
Yongsheng Dong.)}

\thanks{X.~Li is with the Center for OPTical IMagery Analysis and Learning (OPTIMAL), Xi'an Institute of Optics and Precision Mechanics, Chinese Academy of Sciences, Xi'an 710119, Shaanxi, P. R. China (email: xuelong\_li@opt.ac.cn).}
\thanks{K.~Liu is with the Center for OPTical IMagery Analysis and Learning (OPTIMAL), Xi'an Institute of Optics and Precision Mechanics, Chinese Academy of Sciences, Xi'an 710119, Shaanxi, P. R. China (email: kang.liu.opt@gmail.com).}

\thanks{Y. Dong is with the Center for OPTical IMagery Analysis and Learning (OPTIMAL), Xi'an Institute of Optics and Precision Mechanics, Chinese Academy of Sciences, Xi'an 710119, China, and also with the School of Information Engineering, Henan University of Science and Technology, Luoyang 471023, Henan, P. R. China (email: dongyongsheng98@163.com).}
\thanks{D. Tao is with the UBTECH Sydney Artificial Intelligence Institute and the School of Information Technologies in the Faculty of Engineering and Information Technologies at The University of Sydney, J12, 1 Cleveland St, Darlington, NSW 2008, Australia (email: dacheng.tao@sydney.edu.au).}

\thanks{\copyright 20XX IEEE. Personal use of this material is permitted. Permission from IEEE must be obtained for all other uses, in any current or future media, including reprinting/republishing this material for advertising or promotional purposes, creating new collective works, for resale or redistribution to servers or lists, or reuse of any copyrighted component of this work in other works.}

}

\markboth{IEEE Transactions on Neural Networks and Learning Systems}%
{Shell \MakeLowercase{\textit{et al.}}: Bare Demo of IEEEtran.cls for Journals}

\maketitle

\begin{abstract}
Image matting is generally modeled as a space transform from the color space to the alpha space. By estimating the alpha factor of the model, the foreground of an image can be extracted. However, there are some dimensional information redundancy in the alpha space. It usually leads to the misjudgments of some pixels near the boundary between the foreground and background. In this paper, a manifold matting framework named Patch Alignment Manifold Matting (PAMM) is proposed for image matting. In particular, we first propose a part modeling of color space in the local image patch. We then perform whole alignment optimization for approximating the alpha results by using subspace reconstructing error. Furthermore, we utilize the Nesterov's algorithm to solve the optimization problem. Finally, we apply some manifold learning methods in the framework, and obtain several image matting methods, such as named ISOMAP matting and its derived Cascade ISOMAP matting (CasISO matting). The experimental results reveal that the manifold matting framework and its two examples are effective when compared with several representative matting methods.
\end{abstract}


\IEEEpeerreviewmaketitle

\section{Introduction}
\label{sec:1}
\IEEEPARstart{I}{mage} matting is an important but still challenging problem in the field of image processing and computer vision, and it is usually implemented by extracting the foreground of an image in terms of estimating a proportional factor which measures the degree of a pixel belonging to the foreground. It has been widely used in image compositing, video editing, film production, and so on. For example, image matting techniques can be used to create image composition or promote further editing tasks \cite{chen2015region,Alpha2000estimation,zhao2015real,Group2014Sparse,gong2015integrated,Lazy2014Superpixel,zhang2015mtc,liu2015random,DBLPYangGTLL}, and to extract the moving objects in the video and re-composite them into desired scenes \cite{Shen2016,wang2016tracking,zhang2015learning,duan2015compressive,wang2014tracking,zhang2016image,Fuyun2008Locality,maksai2016players}.

Image matting is different from saliency detection, though they are much alike. Saliency detection is a computational process to predict such salient stimuli (regions) in images or videos for humans, and has seen many applications in content-aware image editing, adaptive image/video, displaying, and advertisement \cite{lang2016dual}. While image matting particularly pays attention to the boundaries between the foreground and background, and its main task is how to make the boundaries accurate enough.

To properly extract semantically meaningful foreground objects, users usually manually label the foreground, background and unknown regions of an input image before matting. This three parts form the trimaps as shown in Figure \ref{fig1}. Using the trimaps, the problem of image matting then turns into estimating the alpha values for the pixels in unknown regions based on the known foreground and background pixels. Except for the trimaps, another type of prior information can be obtained by the strokes which demand fewer user interaction and operation. The strokes-based algorithms consider the marked scribbles as the input to extract the alpha matte. In terms of application purposes, the trimaps method is appropriate for the matting situations where high quality matting is demanded, while the strokes method is suitable for matting cases where no high accuracy matting is required, but free-style user interaction is preferred \cite{zhu2013s}.
With the development of computer science and digital imaging technologies, the image matting is drawing more and more attention from both professionals and consumers. A variety of image matting methods have been proposed in the past decades. The current typical image matting algorithms can be categorized into sample-based matting methods and affinity-based matting methods.
\par 
Sample-based matting methods \cite{karacan2015image,Matting07Survey,Targeting2014Object,Optimized2007color} usually consider one pixel is surrounded by a local region and perform image matting by sampling some neighbor pixels in certain rules. Among them, Bayesian matting \cite{A2001bayesian}, and Shared matting \cite{Shared2010} are two typical methods. The Bayesian matting is based on the Ruson and Tomasi' algorithm and uses a continuously sliding small window for the neighborhood definition to model Gaussian mixtures for the foreground and background. Shared matting assumes that the neighborhood pixels share similar attributes in a small observation window, and aiming at the real-time matting technique. Sample-based matting methods perform well for some simple color patterns. However, this type of methods cannot process the complex object effectively, because the matting results particularly rely on the sampling strategy.

\par 
For avoiding the disadvantage of the sample-based matting method, affinity-based matting methods \cite{JSun2010fast,Constrained2014Laplacian,TV2012,Regularized2012Matte} implement image matting by the assumption of local smoothness. This is because the correlation of foreground pixels and background pixels is more strong in a small local window. For example, Poisson matting \cite{Poisson2004matting} assumes the background and the foreground colors are locally smooth for unknown pixels, then it needs to solve a homogenous Laplacian matrix. Technically, it uses an approximate gradient field of mattes. Closed-form matting approach \cite{Closed2008Form} is proposed by introducing the matting Laplacian matrix and solving a quadratic cost function under the assumption that the colors, both foreground and background, can be fit with linear model in the local window. For Spectral matting \cite{Spectral2008matting}, there is a much important conclusion that the smallest eigenvectors of the matting Laplacian matrix span the individual matting components of the image, thus the image mask can be recovered by these components linearly. Towards the images with complex textures, affinity-based matting methods performs more robustly than that of sample-based matting methods. However, the computation complexity of affinity-based matting methods is mostly high. This is because this type matting methods usually need to solve a huge affinity matrix. Moreover, these methods may dim the definite boundary.

\begin{table*}[!ht]\footnotesize
\caption{Important Notations Used in the Paper.}\label{tab0}
\begin{center}
\begin{tabular}							
{|c|c||c|c|} \hline
\emph{\textbf{Notation}} & \emph{\textbf{Description}} & \emph{\textbf{Notation}} & \emph{\textbf{Description}}  \\ \hline
{$X$} & given dataset in a high dimensional space & $L_i$ & representation of $i$-th patch \\ \hline
{$Y$} & obtain dataset in a reduced lowly space  & $S_i$ & selection matrix \\ \hline
{$I_i$} & color of $i$-th pixel & $w_i$ & local window for $i$-th pixel \\ \hline
{$I_{i_j}$} & color of $j$-th neighbor of $i$-th pixel & {$\xi _{j}^{(i)}$} & reconstruction error of $j$-th neighbor of $i$-th color patch \\ \hline
{$X_i$} & patch $i$ in a high dimensional space & {$\varepsilon_j^{(i)}$} & reconstruction error of $j$-th neighbor of $i$-th alpha patch \\ \hline
{$Y_i$} & patch $i$ in a reduced lowly space  & $Y_{i}^{+}$ & Moor-Penrose pseudo inverse of $Y_{i}$ \\ \hline
{$N$ and $p$} & number of pixels in a local patch & ${W_{ij}}$ & weighted factor between ${x_i}$ and its neighbor $j$ \\ \hline
{$P_i$} & affine transformation between $Y_i$ and $X_i$ & {$E$} & identity matrix \\ \hline
{$Q_i$} & $d$ orthonormal columns & $k$ & number of neighbors \\ \hline
{${{\alpha }_{i}}$} & matting feature of $i$-th pixel & $b$ & a known vector from the trimaps \\ \hline
${A _i}$ & alpha vector in $i$-th patch & $\left\{ {{A_k}} \right\}$ & the sequence of approximate solutions \\ \hline
{$E_i$} & reconstruction error of $i$-th patch in alpha space & $\left\{ {{s_k}} \right\}$ & the sequence of search points \\ \hline
$e$ & a vector with $p$ dimensions & ${\beta _k}$ & a variable as the iterations \\ \hline
\end{tabular}
\end{center}
\end{table*}

\par 
To make good use of the idea of sampling and affinity, some confusion methods \cite{Grabcut2004,Random2005walks,Easy2006matting,Robust2010,A2007geodesic} are proposed to solve the image matting problem. Actually, it is a trade-off due to that the confusion matting methods are composited by sample-based matting method and affinity-based matting methods. However, they still can not entirely alleviate the problems confronted by the two types of matting methods. As such, much more methods are proposed for image/video matting \cite{Jubin2016Sparse,Zou2015Video,Nonlinear2014Seg,kang2015Superpixel,alush2016hierarchical}. More typically, the work \cite{fiss2015light} is based on the prior information of light field, and \cite{zhou2014automatic} is based on defocus spectral information. For further capturing the low-dimensional manifold structure of complex pattern, a few manifold matting methods are proposed in recent years. They include LTSA matting \cite{LTSA2011}, LLE matting \cite{LLE2013matting}, and \cite{Unsupervised2004,DBLP09SubspaceSelection,Db2006Unsuper,wang2011subspaces}. These manifold matting algorithms are designed by optimizing some intuitive energy functions.
In fact, there are a multitude of manifold learning methods, including Laplacian Eigenmaps (LE) \cite{LE2003}, Maximum Variance Unfolding (MVU) \cite{DBLP_conver13MVU}, ISOMAP \cite{ISOMAP2000}, Locality Preserving Projections (LPP) \cite{he2003locality}, and so on \cite{zhang2015Ensemble,zhang2013Tensor}. Each of them has its own advantage in processing high-dimensional data. But the aforementioned manifold matting algorithms can not be effectively fused by the manifold learning methods.

Motivated by the above problems, we in this paper propose a unified manifold matting framework named as Patch Alignment Manifold Matting (PAMM) for image matting. The idea is that under the assumption that the alpha space shares a common subspace with the color space, we propose a manifold model of local image patches in color space, and attempt to mine the intrinsic information of the alpha space to compute the alpha value by using patch alignment manifold learning. In particular, PAMM mainly consists of part modeling and whole alignment. The part modeling is used to produce the alpha reconstruction error of one local patch, while the whole alignment is performed to derive the whole alpha reconstruction error. Moreover, prior information is offered with trimaps for the optimization problem. This is because the image matting methods need the prior information to learn the definite foreground, definite background, and unknown region.

Furthermore, we utilize an efficient Nesterov's algorithm \cite{Nesterov1983,MarcTeboulle2009,Nesterov2007} to iteratively solve the optimization problem until the users need is met \cite{An2005iterative}. Then the expected alpha mask is obtained and utilized to extract the image foreground. Finally, we construct some concrete example of our proposed PAMM framework. Among them, the two new manifold learning matting algorithms, termed ISOMAP matting and its derived Cascade ISOMAP matting (CasISO matting), are more effective. The experimental results reveal the effectiveness of the manifold matting framework and its two example methods by comparing with several representative matting methods.

\par 
The contributions of this paper are as follows. First, we propose a unified manifold image matting framework called PAMM into which different manifold learning methods can be incorporated to produce the corresponding image matting technologies. Due to that this framework is constructed to adaptively mine the intrinsic information in the alpha space, it can process the complex pattern of an image better than the current representative methods.
Second, we present two efficient implementations, ISOMAP matting and CasISO matting, to manifest the universal application of the framework. This two matting methods can deal with the nonlinear data distribution and well preserve discriminability of pixel classes.
Third, we perform extensive experiments for comparing our proposed methods with eight current representative methods. Experimental results reveal the effectiveness and superiority of our proposed methods.

\par 
The remainder sections of this paper are organized as follows. In section II, Some related works are presented. The proposed matting framework PAMM and some representative manifold matting methods including ISOMAP matting and CasISO matting are described in section III. Experimental results are shown in section IV. Finally, we present the conclusions and future work.

\section{Related Work}

In this section, we will introduce some related works about dimension reduction which is a key step in our proposed manifold matting scheme. Locally Linear Embedding (LLE) \cite{LLE2000} is a powerful eigenvector method for the problem of the nonlinear reduction. The LLE uses linear coefficients, which reconstruct a given measurement by its neighbors, to represent the local geometry. Then the LLE seeks a low dimensional embedding in which these coefficients are still suitable for reconstruction. Therefore, the LLE is an unsupervised learning algorithm which is based on the linear structure over a local window. Local Tangent Space Alignment (LTSA) \cite{Zhang2004LTSA} exploits the local tangent information as a representation of local geometry, and this local tangent information is then aligned to provide a global coordinate. The LTSA matting assumes that the local smoothness assumptions have been replaced by implicit manifold structure defined in local color spaces and formulate a new cost function. The algorithm of LTSA first extracts local information and then constructs alignment matrices, followed by aligning global coordinates. The Laplacian Eigenmaps (LE) \cite{LE2003} is a geometrically motivated approach to nonlinear dimensionality reduction which has locality-preserving properties and natural connections to the graph embedding clustering. This dimensionality reduction method constructs an undirected and weighted graph to describe the data manifold, and the low-dimensional data can be found by solving the graph embedding. The Maximum Variance Unfolding (MVU) \cite{wang2011Maximum}, also called Semidefinite Embedding (SDE), uses Semidefinite Programming (SDP) and Kernel Matrix Factorization (KMF) to model nonlinear dimensionality reduction problems. The ISOMAP \cite{ISOMAP2000} is also a excellent manifold learning method estimating the geodesic distance between faraway points. The ISOMAP preserves global geodesic distances of all the pairs of measurements. For neighboring points, input space distance provides a good approximation to geodesic distance. These approximations are computed efficiently by finding shortest paths in a graph with edges connecting neighboring data points.

\section{Patch Alignment Manifold Matting Framework}
\label{sec:2}
For the problem of image matting, a basic assumption model is mostly used as Eq. (\ref{eq:0}).
\begin{eqnarray}\label{eq:0}
{I_i} = {\alpha _i}{F_i} + (1 - {\alpha _i}){B_i},
\end{eqnarray}
where the color of pixel $i$ can be denoted as $I_i$. The image can be regarded as two components: foreground image $F_i$ and background image $B_i$, and the term $\alpha _i$  means the opacity of the pixel $i$ and balances the two components. This formulation is an illness model with many unknown terms so that the biggest challenge of image matting is how to find the optimal alpha solutions.
In fact, there are some dimensional information redundancy in the alpha space. We can solve the problem by using the patch alignment method.
\par
The idea of patch alignment is firstly introduced in \cite{Patch2009DBLP}. It reveals the intrinsic structure on which most of the nonlinear dimensionality reduction methods and manifold learning methods are based \cite{Fuyun2008Locally,Gao2010,Yu2012Adaptive,Du2014Target,Ding2016}. This framework consists of two parts: part modeling and whole alignment. For part modeling, different algorithms have different optimization criteria over patches, and each of them is built using a certain distance measurement with its related points. The part modeling usually applies manifold learning algorithms, such as LTSA, LLE, MVU, ISOMAP, and so on. For whole alignment, all part modelings are integrated to form final global coordinate for all of the independent patches based on the alignment strategy, originally used in \cite{Zhang2004LTSA}. And the whole alignment stage unifies dimensionality reduction algorithms of spectral-based analysis. This framework discovers that 1) algorithms are intrinsically different in the patch optimization stage, and 2) all algorithms share an almost identical whole alignment stage. Some important notations used in the paper are summaried in Table \ref{tab0}.

For the part modeling, different algorithms have different optimization methods over patches. In the part modeling step, we consider any measurement ${x_i}$  and its $\bm{\emph{\textbf{k}}}$  related nearest neighbors ${x_1,x_2,...}$ and ${x_i}$. The matrix ${X_i} = [{x_i},{x_{{i_1}}}, \ldots ,{x_{{i_k}}}] \in {\rm{I}}{{\rm{R}}^{m \times (k + 1)}}$ is formed to denote the patch. For ${X_i}$, we have a part mapping ${f_i}:{X_i} \mapsto {Y_i}$  and ${Y_i} = [{y_i},{y_{{i_1}}}, \ldots ,{y_{{i_k}}}] \in {\rm{I}}{{\rm{R}}^{d \times (k + 1)}}
$. The part modeling is defined as
\[\arg \mathop {\min }\limits_{{Y_i}} {\rm{tr}}({Y_i}{L_i}Y_i^T) ,\]
where ${\rm{tr}}( \cdot )$ is the trace operator; ${L_i} \in {\rm{I}}{{\rm{R}}^{(k + 1) \times (k + 1)}}$ varies with the different algorithms, encoding the objective function for the $i$-th patch.

In whole alignment stage, all part modelings are integrated to form the final global coordinate for all independent patches. We denote the ${Y_i}$ as a low dimensional data representation for each patch  ${X_i}$. Assuming that the coordinate of each patch ${Y_i}$  is selected from the global coordinate ${Y}$, that is
\[{Y_i} = Y{S_i} ,\]
where ${S_i} \in {\rm{I}}{{\rm{R}}^{N \times (k + 1)}}$  is the selection matrix and an entry is defined as
\[{({S_i})_{bq}} = \{ \begin{array}{*{20}{c}}
   {1,} & {b = {F_i}\{ q\} }  \\
   {0,} & {else}  \\
\end{array} ,\]
where ${F_i} = \{ i,{i_1}, \ldots ,{i_k}\}$ denotes the index set for the patch $i
$ that is consists of the measurement  ${x_i}$ (or ${y_i}$) and its $\bm{\emph{\textbf{k}}}$ related neighbors. Thus, the formulation can be
rewritten as
\[
\arg \mathop {\min }\limits_Y {\rm{tr}}(Y{S_i}{L_i}S_i^T{Y^T}) .\]
Gathering all of the patch ${X_i}$s, there will be a whole alignment matrix. The whole alignment can be derived as
\[ \begin{array}{l}
 \arg \mathop {\min }\limits_Y \sum\limits_{i = 1}^N {{\rm{tr}}(Y{S_i}{L_i}S_i^T{Y^T})}  \\
  = \arg \mathop {\min }\limits_Y {\rm{tr}}(Y(\sum\limits_{i = 1}^N {{S_i}{L_i}S_i^T} ){Y^T}) \\
  = \arg \mathop {\min }\limits_Y {\rm{tr}}(YL{Y^T}) ,\\
 \end{array} \]
where $N$ denotes the number of the image patches; and $L = \sum\limits_{i = 1}^N {{S_i}{L_i}S_i^T \in {\rm{I}}{{\rm{R}}^{N \times N}}}$  is the alignment matrix. This is the idea of patch alignment. In the following subsections, the framework of PAMM will be presented.

\begin{figure*}[!ht]\footnotesize 
\setlength{\abovecaptionskip}{0pt}
\setlength{\belowcaptionskip}{0pt}
\centering
\begin{overpic}[scale=0.5]{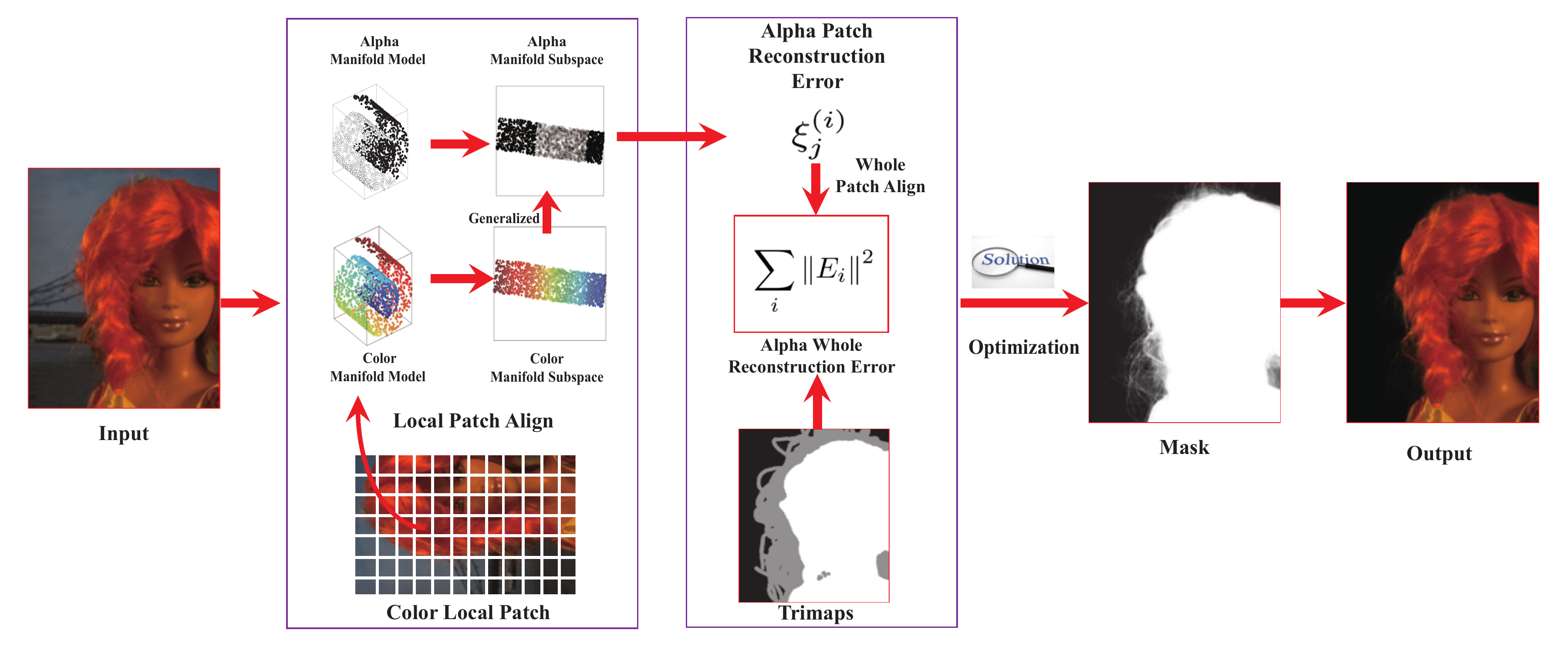}
\end{overpic}
\caption{Flow diagram of the proposed patch alignment manifold matting (PAMM).}\label{fig1}
\end{figure*}

\emph{1) PAMM Part Modeling.}
For the image matting problem, the manifold learning methods are utilized on small image patches. The methods are conducted on the RGB color space to find the subspace, so that we can obtain the reconstruction error between the observation data and the assumption model. Then the reconstruction error can be optimized to minimum energy. In this paper, we try to explore the possibility of applying dimensionality reduction algorithms, particularly manifold learning techniques, to solving image matting problem based on the alpha model. For image matting problem, geometric structure of pixels must be taken into consideration firstly, so small local windows are defined in which the neighbors of a pixel data are chosen from. For pixel $i$ and its RGB color vector $I_i$, the neighborhoods of the $I_i$ are defined as the vectors $I_{i_j}$. And ${i}_{j}$ denotes the $j$-th neighbor of the pixel $i$ in terms of a local window ${{w}_{i}}=\{{{i}_{1}},...,{{i}_{p}}\}$, usually a $3\times 3$ window with $i$ as the center pixel. For each pixel $i$, define a subset
\begin{eqnarray}
{{X}_{i}}=\{{{I}_{{{i}_{j}}}}|{{i}_{j}}\in {{w}_{i}},j=1,...,p\} ,
\end{eqnarray}
where $p$ is the pixel index in the local window.
\par
For the purpose of image matting, there is a basic assumption that the pixel of natural color image has three alpha channels which correspond to the RGB color channels. The affinity subspace of the color space is transformed into that of alpha space. That is to say, the color channels and the alpha channels will share the same color subspace.

The local affine is computed by RGB color vector and its manifold subspace, and the data distribution of a local window is assumed manifold structure. Therefore, in a small local window ${{w}_{i}}=\{{{i}_{1}},...,{{i}_{p}}\}$, the color affine subspace can be used to derive the reconstruction error function of the alpha data space.
The matting problem is aiming at to solve alpha solution of image pixels, so we build a reconstruction error which is based on the color subspace reduced dimensions from the original color data. There is a method which can be utilized to get the affine subspace approximation between the ${{X}_{i}}$ and the ${{Y}_{i}}$ for most of the manifold learning algorithms
\begin{eqnarray}
Y_{i}={P_i}X_{i} ,
\end{eqnarray}
where the ${P_i}$ is the affine transformation which can find the low dimension subspace ${{Y}_{i}}$ of the high dimension data ${{X}_{i}}$.

Applying some manifold learning algorithm over color ${{I}_{i}}$, there will be a ${{Q}_{i}}$ of $d$ orthonormal columns \cite{LTSA2011} such that
\begin{eqnarray}
{{I}_{{{i}_{j}}}}={{\overline{I}}_{i}}+{{Q}_{i}}Y_{j}^{(i)}+\xi _{j}^{(i)} ,
\end{eqnarray}
where $\xi _{j}^{(i)}=(E-{{Q}_{i}}Q_{i}^{T})({{I}_{{{i}_{j}}}}-{{\overline{I}}_{i}})$ with the identity matrix $E$ is the reconstruction error, $Y_{j}^{(i)}$ is the local coordinates over the subspace in the color space and ${{\overline{I}}_{i}}$ is the mean color vector.

For the purpose of image matting, the global matting feature ${{\alpha }_{i}}$ of the local coordinates $Y_{j}^{(i)}$ is reconstructed. This local coordinates are based on the local information on the local manifold defined by the windows. Specifically, we wish for the matting values ${{\alpha }_{{{i}_{j}}}}$ to satisfy the set of the equations as follows, according to local structures determined by the $Y_{j}^{(i)}$,
\begin{eqnarray}
{\alpha _{{i_j}}} = \overline {{\alpha _i}}  + {P_i}Y_j^{(i)} + \varepsilon _j^{(i)},j = 1, \ldots ,p;i = 1, \ldots ,N ,
\end{eqnarray}
\begin{eqnarray}
{A_i} = \frac{1}{p}{A_i}e{e^T} + {P_i}{Y_i} + {E_i},
\end{eqnarray}
where $\overline {{\alpha _i}}$ is the mean of $\overline {{\alpha _{{i_j}}}} (j = 1, \cdots ,p)$, and ${P_i}$ is the local affine transformation matrix in the alpha space. Denote ${A _i} = ({\alpha _{{i_1}}}, \cdots ,{\alpha _{{i_p}}})$, $e=(1,...,1)^{T}$ is a vector with $p$ dimensions, ${Y_i} = (Y_1^{(i)}, \cdots ,Y_p^{(i)})$ and  ${E_i} = (\varepsilon _1^{(i)}, \cdots ,\varepsilon _p^{(i)})$. However, we assume that the color space and the alpha space share the same low dimensional subspace ${Y_i}$.

\emph{2) PAMM Whole Alignment.}
In the whole alignment, the ${{Y}_{i}}$ is represented as the low dimensional data for each patch ${{X}_{i}}$. Combining all of the unknown reconstruction errors of image patches, we can derive a total reconstruction error
\begin{eqnarray}
\underset{{P_i},{{\alpha }_{i}}}{\mathop{\min }}\,\sum\limits_{i}{{{\left\| {{E}_{i}} \right\|}^{2}}}=\sum\limits_{i}{{{\left\| {{A}_{i}}(E-\frac{1}{p}e{{e}^{T}})-{P_i}{{Y}_{i}} \right\|}^{2}}} ,
\end{eqnarray}
where ${{Y}_{i}}$ is known and ${P_i}$ is the affine transformation matrix. Then we can obtain ${{W}_{i}}=(E-\frac{1}{p}e{{e}^{T}})(E-Y_{i}^{+}Y)$ by using the method in \cite{LTSA2011,Patch2009DBLP}, where $Y_{i}^{+}$ is the Moor-Penrose pseudo inverse of $Y_{i}$, and ${{E}_{i}}={{A}_{i}}{{W}_{i}}$, hence
\begin{eqnarray}
\sum\limits_{i}{{{\left\| {{E}_{i}} \right\|}^{2}}}=\sum\limits_{i}{{{\left\| {{A}_{i}}{{W}_{i}} \right\|}^{2}}} ,
\end{eqnarray}
Note that the components consist of ${{A}_{i}}$ patches and are overlapped, and thus this formula can be rewritten as
\begin{eqnarray}
\sum\limits_{i}{{{\left\| {{E}_{i}} \right\|}^{2}}}=ASW{{W}^{T}}{{S}^{T}}{{A}^{T}}=AM{{A}^{T}} ,
\end{eqnarray}
where $M=SW{{W}^{T}}{{S}^{T}}$, $S=[{{S}_{1}},\ldots ,{{S}_{N}}]$ is the selection matrix and $W=\text{diag(}{{W}_{1}},\ldots ,{{W}_{N}})$. The matrix $M$ is called the patch alignment matrix. The manifold learning methods will share the same framework proposed, and they respectively have different patch alignment matrixes with their own subspace error.

Note that in the assumption model of Eq. (\ref{eq:0}), ${\alpha }_{i}$ means the opacity of the pixel $i$, which is used to balance the foreground image and background image. We construct the vector ${A _i}$ to represent the vector consisting of ${\alpha }_{i}$s of the pixel $i$ with its $p-1$ neighbors. In this way, image matting can be reduced to the problem of solving ${\alpha }_{i}$ or ${A _i}$. By using our proposed manifold matting framework, we can compute the alpha value corresponding to each pixel, and thus each pixel can be classified as a foreground one or a background one.

\par
The proposed PAMM scheme is summarized in Figure \ref{fig1}. We assume that the color space (RGB) can be approximated by the manifold subspace in local patch. The manifold learning methods, $i.e.$ LTSA, LLE, MVU, ISOMAP, are utilized to calculate color subspace for the color space. Besides, the alpha space is assumed to have the shared data subspace with its color space. Applying the color subspace, the alpha patch reconstruction error is obtained. Then aligning the reconstruction errors of all the patches, the energy optimization is derived. The final alpha solution of the energy function will be optimized using iterative shrinkage-thresholding algorithms. After solving the problem, the foreground $F$ and background $B$ are reconstructed using the $\alpha$. Its procedure is presented in Algorithm \ref{alg:Framwork0}. In the following subsections, we will present several derived manifold matting methods based on the PAMM framework.

\par
With a linear term, the problem of total error above will be derived as a function
\begin{eqnarray}
\label{eq1}
\mathop {{\rm{min}}}\limits_{{\rm{0}} \le A \le 1} {\rm{ }}f(A) = AM{A^T} + b{A^T} ,
\end{eqnarray}
where $b$ is a known vector from the trimaps, and it has the same dimension as $A$. This function is still a smooth and
convex problem.
\par
In order to determine $A$, we apply the Nesterov's algorithm \cite{Nesterov1983,MarcTeboulle2009,Nesterov2007} which has been proved as an optimal first order method for smooth convex optimization to solve this problem. As same as the gradient method, the Nesterov's algorithm does not require more than one gradient evaluation at each iteration, but just an additional point that is smartly chosen and easy to compute. Besides, the convergence rate of this optimization algorithm is with an complexity $O(1/k^2)$. Applying this optimization algorithm, the key steps will be briefly introduced as below.
\par
The optimization function (\ref{eq2}) is modeled for approximating the function $f(A)$ in (\ref{eq1}) at the point $A$
\begin{eqnarray}
\label{eq2}
{h_{C,A}}(\widetilde A) = f(A)+ {f^{'}}{(\widetilde A - A)^T} + \frac{C}{2}{\left\| {\widetilde A - A} \right\|^2} ,
\end{eqnarray}
where $C > 0$ is a constant. In this step, the Nesterov's method is utilized and based on two sequences $\left\{ {{A_k}} \right\}$  which is the sequence of approximate solutions and $\left\{ {{s_k}} \right\}$ which is the sequence of search points that
\begin{eqnarray}
{s_k} = {A_k} + {\beta _k}({A_k} - {A_{k - 1}}) ,
\end{eqnarray}
where ${\beta _k}$ is a coefficient which need to be chosen, whereas it is a variable as the iterations, not a constant.

Then the alpha solution can be solved by the formula as below
\begin{eqnarray}
{A_{k + 1}} = \max \left\{ {0,\min \left\{ {1,{s_k} - \frac{1}{{{C_k}}}{f^{'}}({s_k})} \right\}} \right\} ,
\end{eqnarray}
where ${C_k}$ is determined by line search rule, and the min and max operators are over vectors as well as in Matlab.

\begin{algorithm}[!ht]
\caption{Patch Alignment Manifold Matting}
\label{alg:Framwork0}
\begin{algorithmic}[1]
\REQUIRE
Original image and trimaps

\ENSURE Image mask ${\alpha}$ \\

\STATE Given an image, define a local neighborhood window $w_i$ for each pixel $i$;
\STATE Assume that the color space (RGB) can be approximated by manifold subspace in local patch; and utilize manifold learning methods, $i.e.$ LTSA, LLE, MVU, ISOMAP, calculate color low dimensional data ${Y_i}$ with transformation matrix ${P_i}$ from color space to the color subspace;
\STATE For pixel $i$, the alpha space is assumed to have the same low dimensional data ${Y_i}$ with its color space. Applying ${Y_i}$, obtain the alpha patch reconstruction error $\varepsilon _j^{(i)}$;
\STATE Align all the image patches and formulate alpha whole reconstruction error $E_i$, and then derive the energy optimization formula $\sum\limits_{i}{{{\left\| {{E}_{i}} \right\|}^{2}}}$;
\STATE Iteratively using Nesterov's algorithm with the priori information of trimaps, until the energy is minimum;
\STATE Get the final ${\alpha}$ solution of the image mask.
\end{algorithmic}
\end{algorithm}

\subsection{LLE Matting}
\par
As for the LLE matting \cite{LLE2013matting} of an image, the nearest neighbors are defined from spatial distance. The ${X_i} = \{ {I_{{i_j}}}{\rm{|}}{i_j} \in {w_i},j = 1,...,k\}$ is used to denote the subset of color vectors over local patch window pixels of the $i$-th pixel. Note that the pixel color ${I_i}$ is contained in patch ${X_i}$. Under this LLE assumption, the color vector ${I_i}$ at pixel $i$ can be approximated by a linear combination ${w_{ij}}$ (so-called reconstruction weights) of its $k$-nearest neighbors of ${I_i}$ in patch ${X_i}$. Therefore, LLE fits a hyperplane through $I_i$ and its nearest neighbors in the color manifold space are defined over the image pixels.

For reasonable alpha solutions of image matting, LLE assumes that the local alpha space is preserved well as same as the color manifold. Once the $W = {\rm{\{ }}{w_{ij}}{\rm{|}}i = 1,...,N,j = 1,...,k{\rm{\} }}$ is determined, applying the manifold matting framework, the reconstruction error functions of the LLE matting can be determined by minimizing the following objective function
\begin{eqnarray}
\mathop {\min }\limits_A F(A) = \mathop {\min }\limits_W \sum\limits_{i = 1}^{} {{{\left\| {{\alpha _i} - \sum\limits_{j = 1}^k {{w_{ij}}{\alpha _{{i_j}}}} } \right\|}^2}}  = {A^T}{M_{LLE}}A ,\end{eqnarray}
where $A = \{ {\alpha _1},{\alpha _2},...,{\alpha _N}\}$, ${M_{LLE}} = {(E - W)^T}(E - W)$ is called LLE alignment matrix, and $E$ is the identity matrix.

\subsection{LTSA Matting}
In terms of LTSA matting \cite{LTSA2011}, we define neighborhood points by incorporating pixel geometric structure. For each pixel $i$, we define the neighborhood of $I_i$ as the RGB vector ${{I}_{{{i}_{j}}}}$. And the ${{i}_{j}}$ means pixel $j$ is the neighbor of pixel $i$ in a local window.
Applying the classical PCA over local window $X_i$, there is a ${{Q}_{i}}$ of $d$ (chosen to $< 3$) orthonormal columns such that
${{I}_{{{i}_{j}}}}={{\overline{I}}_{i}}+{{Q}_{i}}Y_{j}^{(i)}+\xi _{j}^{(i)}$.
From above step, we can get the subspace data $Y_{j}^{(i)}$ on the image patch $i$ of the LTSA matting method. Therefore, applying the manifold matting framework, the whole alignment reconstruction error of LTSA matting is as follows
\begin{eqnarray}
\sum\limits_{i}{{{\left\| {{E}_{i}} \right\|}^{2}}}=ASW{{W}^{T}}{{S}^{T}}{{A}^{T}}=A{{M}_{LTSA}}{{A}^{T}} ,
\end{eqnarray}
where ${{M}_{LTSA}}$ is the alignment matrix of LTSA matting method.

\subsection{ISOMAP Matting}
The ISOMAP \cite{ISOMAP2000} is an excellent manifold learning method estimating the geodesic distance between faraway points. So we firstly propose ISOMAP matting method applying ISOMAP on the PAMM framework. This matting method can deal with the nonlinear data distribution and better preserve discriminability of pixel classes.
For the image matting problem, the manifold learning methods are utilized on small image patches. The methods are conducted on the RGB color space to find the subspace, so that we can obtain the reconstruction error between the observation data and the assumption model. Then the reconstruction error can be optimized to minimum energy. 

The ISOMAP method needn't compute the dimensionality reduction over the whole image because there will be much computation cost when the number of data points is exponentially growing. For image matting, it is reasonable to apply ISOMAP method to obtain color subspace in local patches.
Applying the ISOMAP method over the patch $\bm{\emph{\textbf{X}}}_i$ which has three dimension RGB color channels, there will be a subspace $\bm{\emph{\textbf{Y}}}$ of lower dimension. Then we can get the affine subspace approximation from between the $\bm{\emph{\textbf{X}}}_i$ and the $\bm{\emph{\textbf{Y}}}_i$. The formulations can be derived as follows

\begin{eqnarray}
Y_{i}={P_i}X_{i} ,
\end{eqnarray}
where the ${P_i}$ is the affine approximation which can find the low dimension subspace $\bm{\emph{\textbf{Y}}}_i$ of the high dimension data $\bm{\emph{\textbf{X}}}_i$.
Just like the approach above, the error function can be derived as follows
\begin{eqnarray}
\sum\limits_{i}{{{\left\| {{E}_{i}} \right\|}^{2}}}=ASW{{W}^{T}}{{S}^{T}}{{A}^{T}}=A{{M}_{ISO}}{{A}^{T}} ,
\end{eqnarray}
where the matrix ${{M}_{ISO}}$ is called the ISOMAP patch alignment matrix.

\subsection{CasISO Matting}
Based on the ISOMAP matting, we also propose a Cascade ISOMAP matting (CasISO matting). Because we want to try our best to explore the best approximate subspace which can obtain best foreground mask and the minimum of reconstruction error. Compared to the ISOMAP matting, the CasISO matting consists of two stages to find its approximate alpha subspace. In a general way, we in the first stage utilize the manifold learning method ISOMAP to transform the color space of an image into one color subspace. Using the same ISOMAP method, we then transform this color subspace into another color subspace in the second stage. This color subspace will be shared to the alpha space. Note that the data space structure will be fully adjusted, and the color subspace will be better for the alpha subspace in this strategy.

\subsection{Other Examples}
Using the same strategy, the LE matting can obtain the reconstruction error function as below
\begin{eqnarray}
\sum\limits_i {{{\left\| {{E_i}} \right\|}^2}}  = ASW{W^T}{S^T}{A^T} = A{M_{LE}}{A^T} ,
\end{eqnarray}
where the ${M_{LE}}$ is the patch alignment matrix of the LE matting method.
And also for the MVU matting, we can get the whole alignment reconstruction error
\begin{eqnarray}
\sum\limits_{i}{{{\left\| {{E}_{i}} \right\|}^{2}}}=ASW{{W}^{T}}{{S}^{T}}{{A}^{T}}=A{{M}_{MVU}}{{A}^{T}} ,
\end{eqnarray}
where the matrix ${{M}_{MVU}}$ is called MVU patch alignment matrix.
\par
These manifold learning methods are good at finding the shared subspace of the color space and the alpha space, and then they are in favor of deriving the whole reconstruction error. Finally, the optimized energy of the whole reconstruction error will be solved by the Nesterov's algorithm. Hence, these manifold learning methods are good fit for the PAMM.

\section{Experiments}
\label{sec:4}
In this section, we will demonstrate the effectiveness of  the ISOMAP matting and CasISO matting methods resulted from our proposed manifold matting framework PAMM  by comparing them with the current representative matting methods.

\subsection{Experimental Settings}
\subsubsection{Dataset}

\begin{figure*}[!ht]\footnotesize 
\setlength{\abovecaptionskip}{0pt}
\setlength{\belowcaptionskip}{0pt}
\centering
\begin{overpic}[scale=0.54]{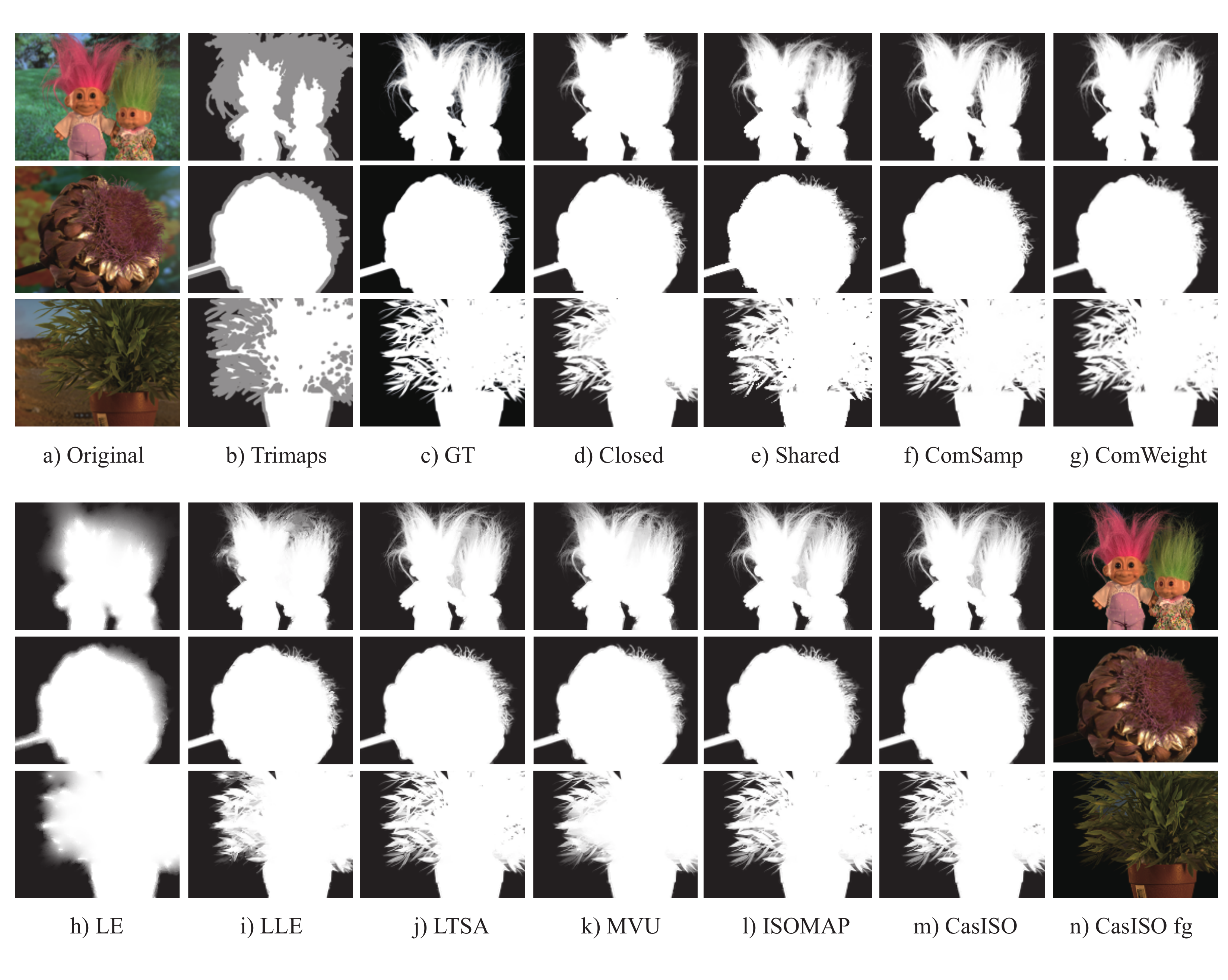}
\end{overpic}
\caption{Matting examples of competitive algorithms. The column a), column b), and column c) are separately original image, trimaps and ground truth image of the three examples. From column d) to column m) are the competitive algorithms including Closed matting \cite{Closed2008Form}, Shared matting \cite{Shared2010}, ComSamp matting \cite{com2013sample}, ComWeight matting\cite{Shahrian2012weighted}, LLE matting \cite{LLE2013matting}, LTSA matting \cite{LTSA2011}, LE matting, MVU matting, ISOMAP matting, and CasISO matting. The column n) shows the foreground image of the three examples by the proposed algorithm CasISO matting.}
\label{fig4}
\end{figure*}

\begin{figure*}
\begin{minipage}{.32\linewidth}
  \centerline{\includegraphics[width=6.3cm]{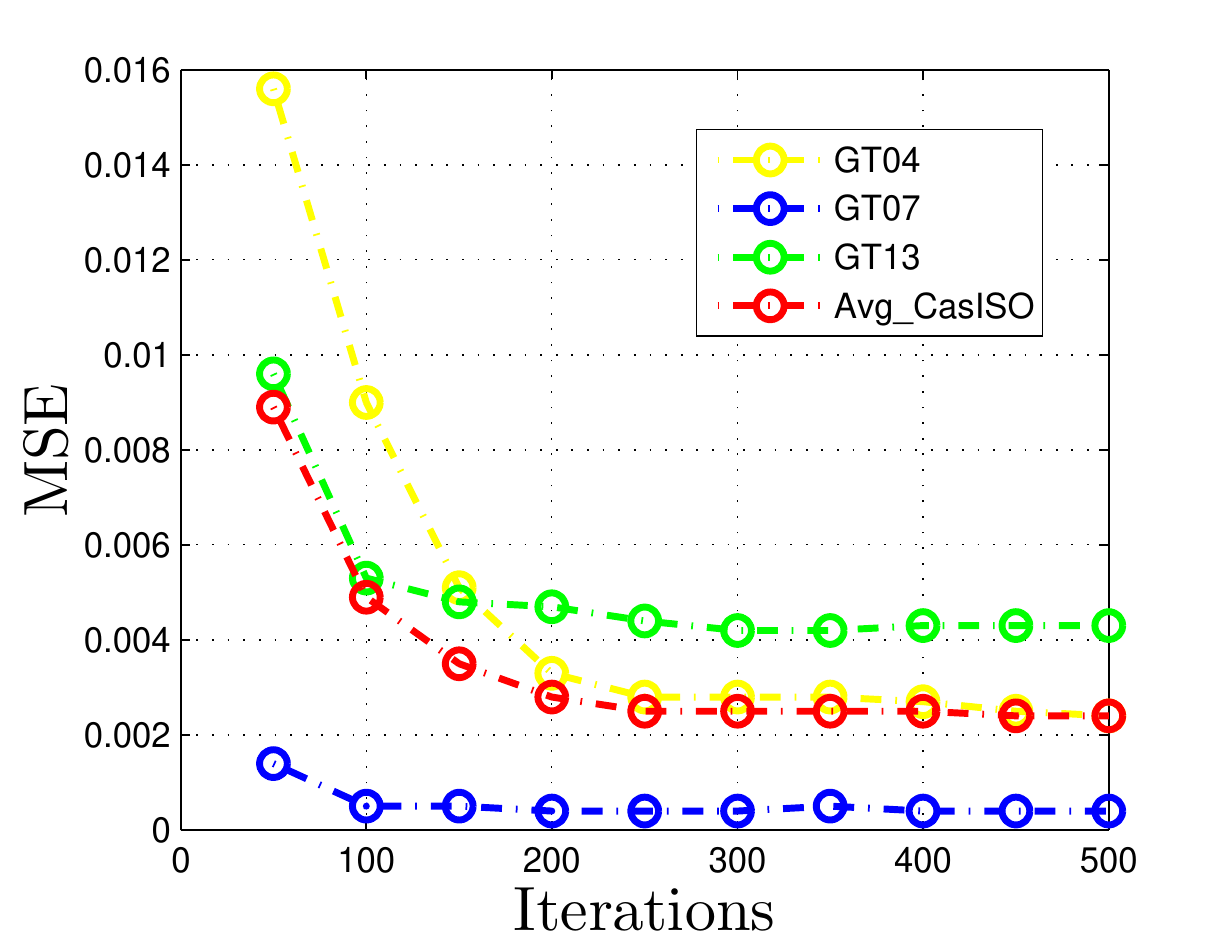}}
  \caption{The average MSE with respect to different K iterations.}
\label{fig3}
\end{minipage}
\hfill
\begin{minipage}{0.32\linewidth}
  \centerline{\includegraphics[width=6.0cm]{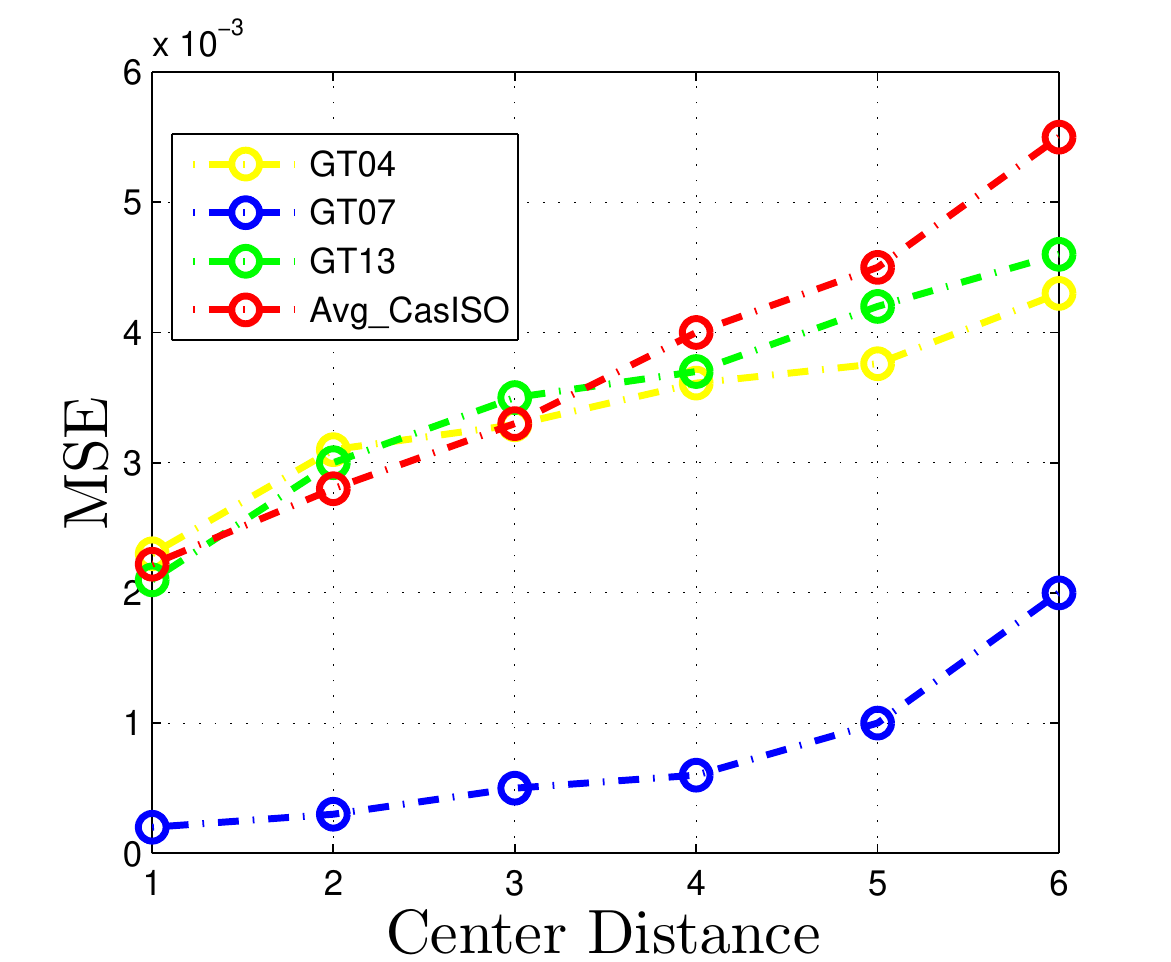}}
  \caption{The average MSE with respect to different center distance of two patches.}
  \label{fig34}
\end{minipage}
\hfill
\begin{minipage}{.32\linewidth}
  \centerline{\includegraphics[width=6.0cm]{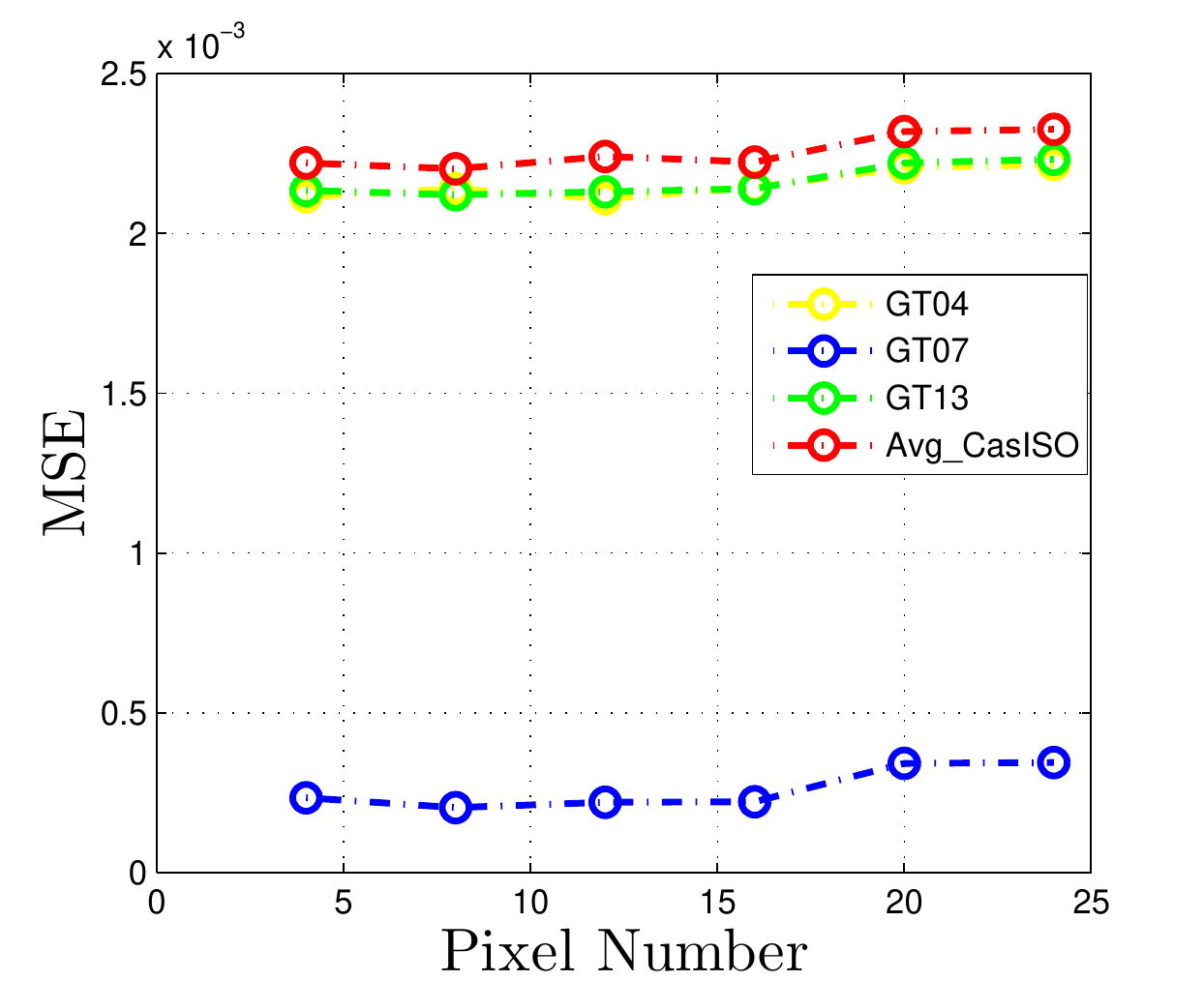}}
  \caption{The average MSE with respect to the pixel number of local patch.}
  \label{fig33}
\end{minipage}
\end{figure*}

\par
In order to demonstrate the effectiveness of the proposed algorithms, one famous publicly available image dataset named alphamatting dataset (http://www.alphamatting.com/) \cite{matting2009Beachmark} is utilized in the experiment. The alphamatting dataset provides testing dataset and training dataset and the ground foreground colors for the images in the training dataset for those who need them. The foreground colors are provided as RGB files. All of the training dataset images are used for the qualitative and quantitative experiment from the dataset. And these images are low resolution training images with some $600 \times 800$ pixels.
There is no existing image matting method which can automatically define the semantic foreground object which fully matches user's requirement. So for image matting, we mostly provide some labels for some pixels with trimaps. In order to obtain a perfect alpha matte result and a fair evaluation, the alphamatting dataset also provides so-called trimaps for the matting algorithms or systems. The trimaps are composed of three parts: definite foreground, definite background and unknown regions.

\begin{figure}[t]
\begin{center}
\includegraphics[width=\linewidth]{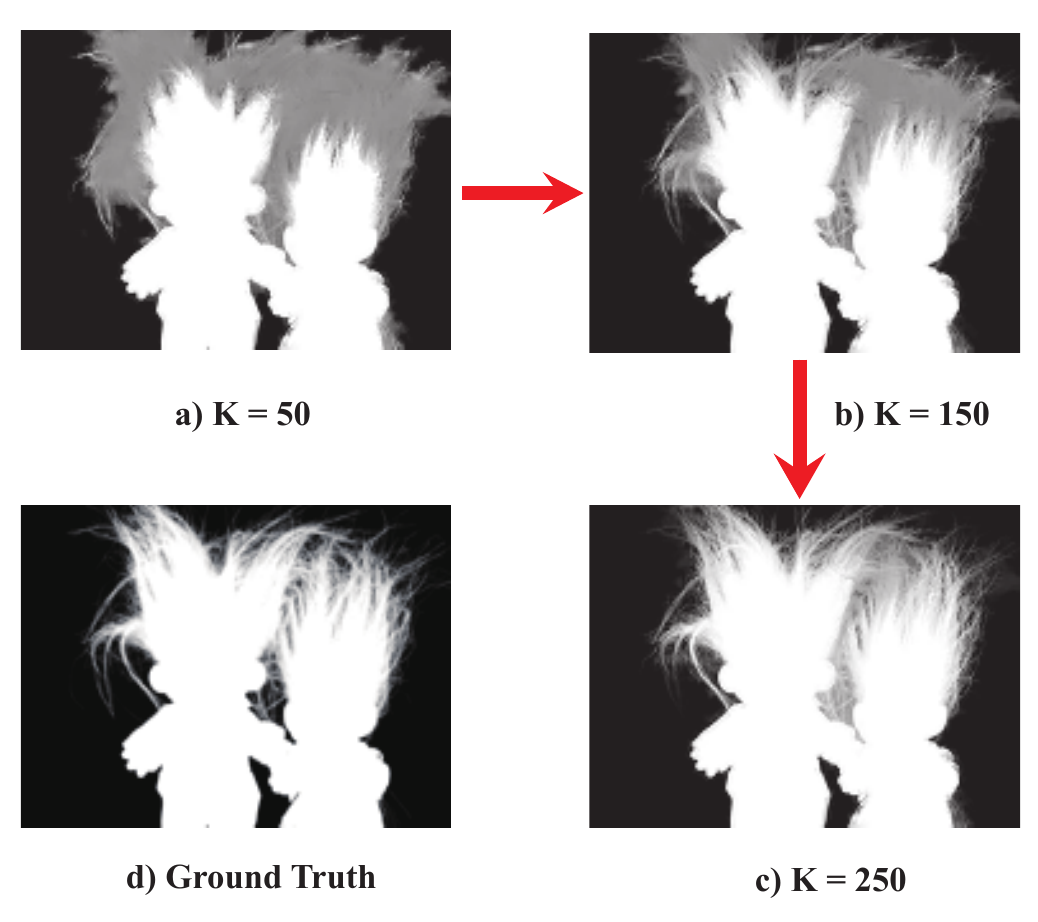}
\end{center}
\vspace{-1em}
\caption{Experimental alpha mask results in different K iterations. The image a), image b) and image c) shows respectively mask result image of CasISO matting method with K = 50, 150 and 250. The image d) shows the mask of ground truth.}
\vspace{-1.5em}
\label{fig2}
\end{figure}

\begin{table}[!ht]\footnotesize
\caption{Average MSE and SAD Results of CasISO matting with different combinations of dimensional reduction.}\label{tab1}
\begin{center}
\begin{tabular}							
{c||c|c|c|c|c|c} \hline Dims & 3-3-2 & 3-3-3 & 3-4-2 & 3-4-3 & 3-5-2 & 3-5-3 \\ \hline
{MSE}
& 0.0025 & \textbf{0.0022} & 0.0025 & 0.0023 & 0.0025 & 0.0023 \\
{SAD}
& 164.93 & \textbf{156.70} & 164.94 & 160.77 & 165.08 & 160.48  \\ \hline
\end{tabular}
\end{center}
\end{table}

\begin{table*}[!ht]\footnotesize
\caption{The Comparisons of MSE Criterion on the Dataset. Compared with other matting methods, the CasISO matting and ISOMAP matting which are based on the proposed manifold matting framework rank first and second.}\label{tab2}
\begin{center}
\begin{tabular}							
{c||c|c|c|c|c|c|c|c|c|c} \hline  & Closed\cite{Closed2008Form} & Shared\cite{Shared2010} & ComSamp\cite{com2013sample} & ComWeight\cite{Shahrian2012weighted} & LLE\cite{LLE2013matting} & LTSA\cite{LTSA2011} & LE & MVU & ISOMAP & CasISO  \\ \hline
{GT01}
& 0.0001 & 0.0004 & 0.0002 & 0.0003  & 0.0002 & 0.0002 & 0.0030& 0.0002 & 0.0002 & 0.0002 \\
{GT02}
& 0.0001 & 0.0018 & 0.0002 & 0.0003  & 0.0003 & 0.0002 & 0.0023& 0.0002 & 0.0002 & 0.0002 \\
{GT03}
& 0.0006 & 0.0005 & 0.0008 & 0.0010  & 0.0001 & 0.0006 & 0.0006& 0.0003 & 0.0004 & 0.0004 \\
{GT04}
& 0.0008 & 0.0025 & 0.0009 & 0.0012  & 0.0048 & 0.0053 & 0.0154& 0.0025 & 0.0020 & 0.0021 \\
{GT05}
& 0.0001 & 0.0008 & 0.0002 & 0.0005  & 0.0003 & 0.0001 & 0.0024& 0.0002 & 0.0001 & 0.0001 \\
{GT06}
& 0.0009 & 0.0012 & 0.0003 & 0.0008  & 0.0010 & 0.0004 & 0.0044& 0.0008 & 0.0005 & 0.0004 \\
{GT07}
& 0.0001 & 0.0004 & 0.0001 & 0.0001  & 0.0004 & 0.0002 & 0.0034& 0.0004 & 0.0002 & 0.0001 \\
{GT08}
& 0.0011 & 0.0028 & 0.0018 & 0.0032  & 0.0011 & 0.0040 & 0.0023& 0.0020 & 0.0015 & 0.0019 \\
{GT09}
& 0.0001 & 0.0006 & 0.0004 & 0.0005  & 0.0011 & 0.0007 & 0.0026& 0.0005 & 0.0004 & 0.0005 \\
{GT10}
& 0.0002 & 0.0018 & 0.0006 & 0.0011  & 0.0014 & 0.0008 & 0.0046& 0.0006 & 0.0006 & 0.0005 \\
{GT11}
& 0.0008 & 0.0024 & 0.0008 & 0.0008  & 0.0012 & 0.0008 & 0.0038& 0.0013 & 0.0009 & 0.0007 \\
{GT12}
& 0.0002 & 0.0008 & 0.0005 & 0.0005  & 0.0003 & 0.0004 & 0.0026& 0.0003 & 0.0003 & 0.0003 \\
{GT13}
& 0.0010 & 0.0045 & 0.0016 & 0.0028  & 0.0088 & 0.0034 & 0.0095& 0.0043 & 0.0022 & 0.0021 \\
{GT14}
& 0.0001 & 0.0005 & 0.0002 & 0.0006  & 0.0002 & 0.0002 & 0.0031& 0.0003 & 0.0002 & 0.0002 \\
{GT15}
& 0.0014 & 0.0029 & 0.0005 & 0.0008  & 0.0013 & 0.0016 & 0.0060& 0.0016 & 0.0017 & 0.0015 \\
{GT16}
& 0.0408 & 0.0354 & 0.0600 & 0.0610  & 0.0308 & 0.0244 & 0.0312& 0.0284 & 0.0281 & 0.0245 \\
{GT17}
& 0.0008 & 0.0014 & 0.0006 & 0.0009  & 0.0004 & 0.0004 & 0.0039& 0.0006 & 0.0004 & 0.0004 \\
{GT18}
& 0.0030 & 0.0014 & 0.0004 & 0.0026  & 0.0017 & 0.0006 & 0.0067& 0.0006 & 0.0005 & 0.0004 \\
{GT19}
& 0.0001 & 0.0008 & 0.0001 & 0.0003  &  0.0004 & 0.0002& 0.0023 & 0.0004 & 0.0002 & 0.0002 \\
{GT20}
& 0.0003 & 0.0007 & 0.0002 & 0.0005  & 0.0005 & 0.0005 & 0.0052& 0.0004 & 0.0003 & 0.0003 \\
{GT21}
& 0.0003 & 0.0022 & 0.0006 & 0.0025  & 0.0018 & 0.0017 & 0.0088& 0.0009 & 0.0010 & 0.0010 \\
{GT22}
& 0.0003 & 0.0007 & 0.0003 & 0.0005  & 0.0004 & 0.0003 & 0.0026& 0.0004 & 0.0003 & 0.0003 \\
{GT23}
& 0.0001 & 0.0008 & 0.0002 & 0.0003  & 0.0002 & 0.0002 & 0.0030& 0.0003 & 0.0002 & 0.0002 \\
{GT24}
& 0.0039 & 0.0011 & 0.0006 & 0.0017  & 0.0006 & 0.0007 & 0.0015& 0.0009 & 0.0008 & 0.0008 \\
{GT25}
& 0.0085 & 0.0130 & 0.0048 & 0.0071  & 0.0069 & 0.0025 & 0.0098& 0.0036 & 0.0025 & 0.0022 \\
{GT26}
& 0.0130 & 0.0156 & 0.0043 & 0.0077  & 0.0218 & 0.0106 & 0.0298& 0.0218 & 0.0128 & 0.0121 \\
{GT27}
& 0.0033 & 0.0113 & 0.0035 & 0.0044  & 0.0120 & 0.0089 & 0.0126& 0.0082 & 0.0059 & 0.0057 \\ \hline
{Avg.}		
& 0.0030 & 0.0040 & 0.0031 & 0.0038  & 0.0037 & \textbf{0.0026} & 0.0068 & 0.0030 & \textbf{0.0024} & \textbf{0.0022} \\ \hline
\end{tabular}
\end{center}
\end{table*}

\subsubsection{Methods for comparisons}
For the proposed manifold matting framework, we can unite different image matting methods to obtain alpha masks. Many manifold learning matting methods, such as LE matting, LLE matting, MVU matting, LTSA matting, ISOMAP matting, and CasISO matting are utilized to make comparisons, both qualitatively and quantitatively. Besides, some other non-manifold newly matting methods are also selected as its competitors. For instance, Closed-Form matting, Shared matting, Com-Sampling matting \cite{com2013sample} and Com-Weighted matting \cite{Shahrian2012weighted} are the latest matting methods which are based on other non-manifold theories. For the structure limit of the table and the figure, the Closed-Form matting, Shared matting, Com-Sampling matting, Com-Weighted matting, LE matting, LLE matting, LTSA matting, MVU matting, ISOMAP matting and CasISO matting in Table \ref{tab2}, Table \ref{tab3} and Figure \ref{fig4} are represented as Closed, Shared, ComSamp, ComWeight, LE, LLE, LTSA, MVU, ISOMAP and CasISO. All of the image matting methods share the same input images and the same trimaps. The manifold image matting methods require large enough memory and long CPU time to iteratively get the final results, so the experimental images in this paper are converted into a certain size with $120 \times 160$ pixels.

\begin{table*}[!ht]\footnotesize
\caption{The Comparisons of SAD Criterion on the Dataset. The CasISO matting and ISOMAP matting which are based on the proposed manifold matting framework rank second and third among the competitive matting methods.}\label{tab3}
\begin{center}
\begin{tabular}							
{c||c|c|c|c|c|c|c|c|c|c} \hline & Closed & Shared & ComSamp & ComWeight & LLE & LTSA & LE & MVU & ISOMAP & CasISO \\ \hline
{GT01}
& 28.38  & 61.07  & 52.18  & 52.28  & 40.54 & 51.18  & 211.67 & 46.61 & 50.22 & 52.62 \\
{GT02}												
& 20.95  & 88.19  & 68.87  & 43.41  & 49.63 & 46.99  & 163.85 & 39.02 & 40.85 & 38.46 \\
{GT03}												
& 135.96 & 120.25 & 176.36 & 182.20 & 51.40 & 58.24  & 82.59  & 84.84 & 79.81 & 79.18 \\
{GT04}												
& 140.08 & 364.94 & 213.45 & 208.77 & 475.87& 542.34 & 969.20 & 383.89 & 337.34 & 360.43 \\
{GT05}												
& 30.97  & 58.27  & 47.89  & 63.57  & 44.82 & 29.13  & 160.20 & 42.44 & 27.84 & 29.21 \\
{GT06}												
& 80.36  & 103.42 & 84.31  & 97.87  & 86.69 & 72.66  & 265.81 & 103.83 & 77.35 & 70.49 \\
{GT07}												
& 16.28  & 46.51  & 27.80  & 29.25  & 62.84 & 45.52  & 237.04 & 61.62 & 40.94 & 37.65 \\
{GT08}												
& 140.01 & 264.65 & 246.29 & 305.43 & 160.66& 342.27 & 222.23 & 252.00 & 228.12 & 251.73 \\
{GT09}												
& 26.29  & 99.73  & 94.95  & 94.88  & 118.24& 103.26 & 205.50 & 89.18 & 89.39 & 102.14 \\
{GT10}												
& 100.61 & 119.63 & 83.95  & 114.13 & 123.67& 103.38 & 274.50 & 89.23 & 84.80 & 79.93 \\
{GT11}												
& 139.72 & 156.18 & 93.42  & 92.73  & 116.93& 95.81  & 291.60 & 127.14 & 99.96 & 94.69 \\
{GT12}												
& 32.84  & 74.62  & 68.65  & 69.00  & 43.88 & 55.59  & 191.13 & 51.03 & 52.35 & 51.08 \\
{GT13}												
& 114.50 & 283.12 & 243.46 & 274.56 & 394.33& 276.41 & 434.97 & 268.65 & 195.60 & 186.87 \\
{GT14}												
& 30.26  & 55.17  & 59.20  & 63.71  & 35.72 & 40.55  & 216.97 & 47.78 & 38.45 & 36.38 \\
{GT15}												
& 131.57 & 144.89 & 65.09  & 89.27  & 116.51& 144.21 & 351.06 & 151.42 & 150.46 & 143.15 \\
{GT16}												
& 1070.44& 968.07 & 1375.41& 1382.19&1042.14& 988.96 & 1105.21& 1051.45 & 1015.16 &	 954.18 \\
{GT17}												
& 104.99 & 124.04 & 102.18 & 123.65 & 73.69 & 87.18  & 247.91 & 98.26 & 87.37 & 84.01 \\
{GT18}												
& 141.98 & 103.94 & 91.42  & 140.28 & 114.34& 79.85  & 360.93 & 89.76 & 75.01 & 68.42 \\
{GT19}												
& 23.36  & 54.04  & 38.72  & 41.89  & 45.22 & 37.24  & 163.08 & 56.76 & 38.46 & 34.02 \\
{GT20}												
& 43.10  & 81.51  & 48.90  & 64.46  & 60.76 & 76.29  & 316.47 & 60.35 & 53.34 & 52.30 \\
{GT21}												
& 42.09  & 182.49 & 169.35 & 203.78 & 165.42& 188.81 & 460.47 & 129.58 & 128.63 & 125.34 \\
{GT22}												
& 60.44  & 87.82  & 75.27  & 84.69  & 68.03 & 78.42  & 218.55 & 83.31 & 72.69 & 74.19 \\
{GT23}												
& 26.26  & 90.12  & 47.84  & 62.18  & 43.49 & 51.56  & 242.54 & 60.31 & 48.08 & 42.70 \\
{GT24}												
& 190.37 & 97.77  & 125.02 & 143.13 & 65.22 & 69.24  & 125.80 & 95.81 & 95.32 & 89.52 \\
{GT25}												
& 282.04 & 453.02 & 239.77 & 334.17 & 309.99& 179.98 & 447.47 & 224.84 & 176.51 & 162.56 \\
{GT26}												
& 515.45 & 714.65 & 432.21 & 520.97 & 791.50& 556.36 & 1070.80& 857.61 & 597.35 & 572.62 \\
{GT27}												
& 198.81 & 497.25 & 280.69 & 318.47 & 497.40& 506.87 & 606.09 & 435.41 & 373.52 & 356.99 \\ \hline
{Avg.}
& \textbf{143.26} & 203.53 & 172.69 & 192.63  & 192.55& 181.79 & 357.17 & 188.23 & \textbf{161.29} & \textbf{156.70} \\ \hline
\end{tabular}
\end{center}
\end{table*}

\subsubsection{Measurements}
Two metric criteria are used to measure the error between the extracted masks and their ground truth. They are \emph{Sum of Absolute Differences} (\textbf{SAD}) and \emph{Mean Squared Error} (\textbf{MSE}), respectively. The \textbf{MSE} is defined as
\begin{eqnarray}
\textbf{MSE} = \frac{1}{{HW}}\sum\limits_{i = 0}^{H - 1} {\sum\limits_{j = 0}^{W - 1} {{{(\textbf{M}(i,j) - \textbf{G}(i,j))}^2}}} ,
\end{eqnarray}
and the \textbf{SAD} is defined as
\begin{eqnarray}
\textbf{SAD} = \sum\limits_{i = 0}^{H - 1} {\sum\limits_{j = 0}^{W - 1} {|\textbf{M}(i,j) - \textbf{G}(i,j)|}} ,
\end{eqnarray}
where $\textbf{M}$ denotes result mask, and $\textbf{G}$ is ground-truth.

\subsection{Matting Performances}

We first investigate the sensitivity of the subspace dimension in the CasISO matting to the matting results. In order to find the best approximate subspace of CasISO matting, several combinations of subspace dimensions are used in our experiments. The experimental results are presented in Table \ref{tab1}. The Dims means variation trend of dimension. First of all, take 3-4-2 for example, the proposed method raises dimension from 3 dimensions to 4 dimensions, and then reduces dimension from 4 dimensions to 2 dimensions. In addition, both of this two steps apply the ISOMAP method. The purpose of raising dimension is to expand the structure of the data, and the dimension reduction wants to find its approximate subspace in nature. From Table. \ref{tab1}, the best combination dims of CasISO matting is obtained by 3-3-3. The MSE of 3-3-3 is with the minimal 0.0022, and the SAD of 3-3-3 is with the minimal 156.70. Although the data dimensions of 3-3-3 are remained, the distribution structure of data space is changed. By this dimensional combination of 3-3-3, the color subspace will be beneficial to the approximation for the original color space and the formation of reconstruction error for the alpha space.

We further perform experiments to find the optimal number of iterations used in the proposed methods. We only test the proposed ISOMAP matting and CasISO matting algorithms due to that they utilize the same efficient Nesterov's algorithm with other example manifold matting algorithms. On the other hand, it is hard to know how many iterations the ISOMAP matting and CasISO matting algorithms need so as to get a optimal alpha mask. Therefore, we compute the MSE criterion of the CasISO matting algorithm in different K iterations as in Figure. \ref{fig3}. Because different image has different convergence iterations, we compute the average MSE on all of the testing image dataset. The red curve in Figure. \ref{fig3} represents the average MSE, and the GT04, GT07 and GT13 are selected randomly to show their differences in convergence iterations. We can see that the average MSE of CasISO matting is becoming stable when the iteration exceeds 250.
Consequently, the iteration of the approximate Nesterov's algorithm for the proposed ISOMAP matting and CasISO matting algorithms is set as 250.

The overlap between patched has been experimented, and showed in Figure \ref{fig34}, the MSE results are growing and the foreground mask are rougher as the center distance of two patches. So the physical meaning of overlap and the experiments demonstrated the overlap is necessary. Usually the local window is with particular size.We have experimented the pixel number in a patch in Figure \ref{fig33}, and the MSE is nearly stable when the pixel number less than 16, while that grows bigger slowly when the pixel number more than 16. However, the pixels in a patch couldn't be too many, because it will be high time and space complexity, and causes bigger MSE reusults. Typically in our PAMM, the size of window $w_i$ is $3\times3$, hence the $p$ is set as 9.

\begin{table*}[!ht]\footnotesize
\caption{The Comparisons of Average E-time on the Dataset.}\label{tab4}
\begin{center}
\begin{tabular}							
{c||c|c|c|c|c|c|c|c|c|c} \hline  & Closed & Shared (C++) & ComSamp & ComWeight & LE & LLE & LTSA & MVU & ISOMAP & CasISO  \\ \hline
{t/seconds}
& 17.42 & 0.37 & 10.61 & 14.63 & 1103.12 & 2147.06 & 935.57 & 6729.19 & 1118.60 & 1324.23 \\ \hline
\end{tabular}
\end{center}
\end{table*}

\begin{figure*}[!ht]\footnotesize 
\setlength{\abovecaptionskip}{0pt}
\setlength{\belowcaptionskip}{0pt}
\centering
\begin{overpic}[scale=0.45]{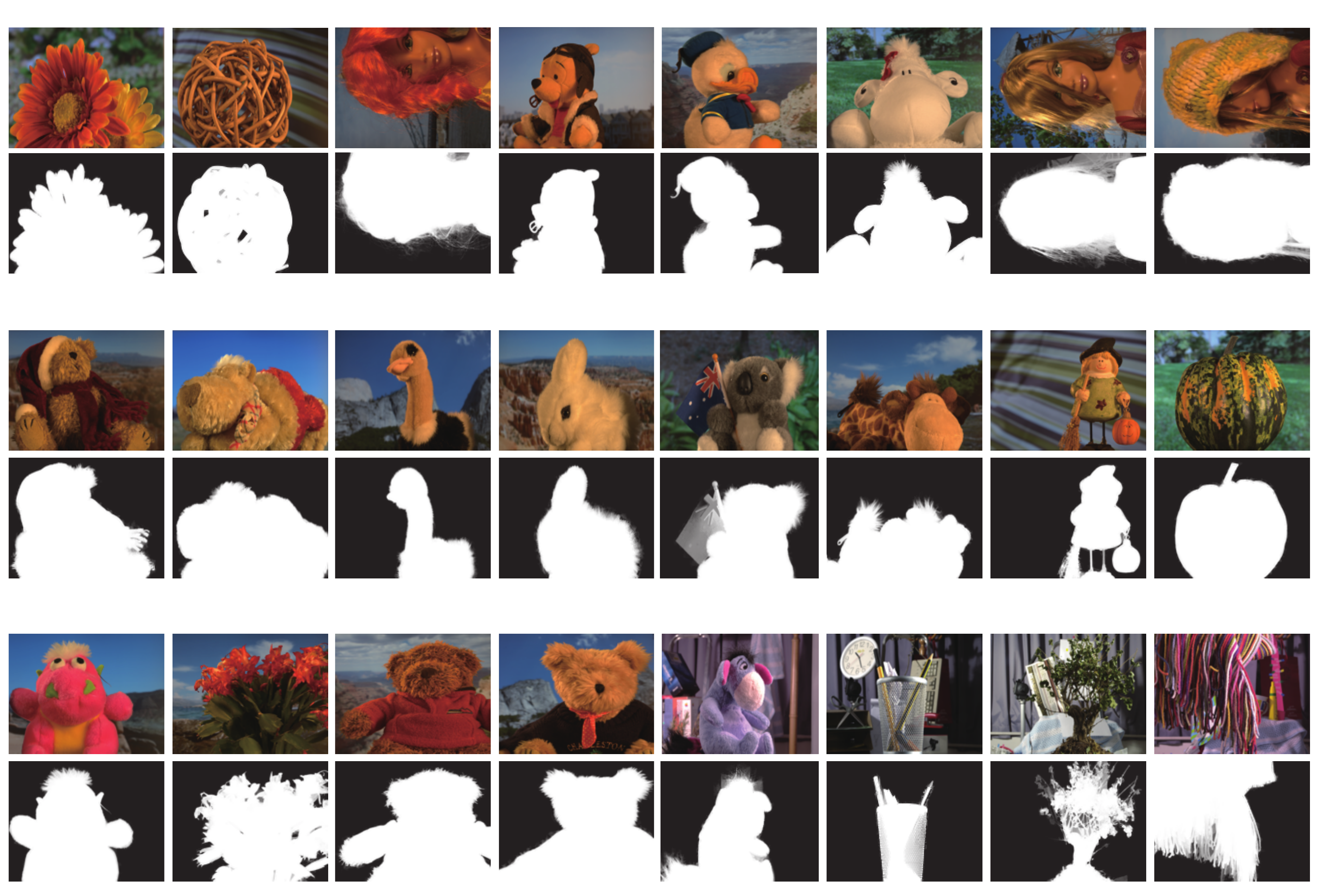}
\end{overpic}
\caption{More matting results of CasISO matting on the alphamatting dataset.}\label{fig5}
\end{figure*}

Experimental alpha mask results in different iterations are also showed in Figure. \ref{fig2}. Obviously, the mask results become more and more accurate as the growing iteration, especially complicated hair in the foreground boundary. Compared to the ground truth image d), the alpha mask image c) with 250 iterations is close to the former in the texture apperance.

\subsection{Comparisons}
\par
In this subsection, the proposed manifold matting framework, ISOMAP matting method, and CasISO matting method are verified both qualitatively and quantitatively on the alpha matting dataset. Three image matting examples of all competitive algorithms are showed qualitatively in Figure. \ref{fig4}. The matting methods are non-manifold from column d) to column g), while they are manifold matting methods column h) to column m). Nearly all of these methods have good performance visually, except for the LE matting method. We can see that all the matting examples of LE matting can not obtain accurate foreground.  Compared to the ground truth, all the alpha mask results of proposed ISOMAP matting and CasISO matting are both perform well. In some cases, non-manifold matting methods perform better, such as example image of Com-Sampling method and Com-Weighted method in the first row. However, because of the complicated hair between foreground and background, it is indispensable to make more comparisons quantitatively.

To further show the effectiveness of our proposed methods, we perform more comparative experiments on the testing dataset. The MSE criterion scores and SAD criterion scores on the testing dataset of all the 10 matting algorithms are showed in Talbe \ref{tab2} and Table \ref{tab3}.
The proposed ISOMAP matting method and CasISO matting method perform best in average MSE criterion with 0.0024 and 0.0022 respectively. Although the Closed-Form matting method ranks first in average SAD criterion with 143.26, the CasISO matting method and ISOMAP matting also have more remarkable average SAD scores than others. The CasISO matting method ranks second with 156.70 average SAD score and ISOMAP matting ranks third with 161.29 average SAD score.
For different images, the competitive methods have different performances in Talbe \ref{tab2} and Table \ref{tab3}. For a single image, the proposed ISOMAP matting and CasISO matting obtain nearly all the best results among manifold matting methods, including the LE matting, LLE matting, MVU matting, LTSA matting. Specifically for image GT16, the proposed CasISO matting get the best SAD score with the 954.18 and nearly best MSE score with the 0.0245. Some non-manifold matting methods, such as the Closed-Form matting, Com-Sampling matting and Com-Weighted matting, also obtain some best MSE and SAD results.
However, the data distributions of some examples are very complex and there are some particular cases which are very hard to completely model. Also it is very hard to visually distinguish the differences of masks between CasISO matting and ISOMAP matting. In Table \ref{tab3}, the SAD scores of ISOMAP matting are better for some pictures, while those of CasISO matting outperform better in other cases. However, the CasISO matting outperforms the ISOMAP matting on the whole, and it obtains most of the best MSE and SAD results. Because the manifold data structure of CasISO matting is fully adjusted and the alpha results are much closer to the ground truth. But the time complex of CasISO matting is higher than the ISOMAP matting. We showed the E-time of all comparison methods in Table \ref{tab4}, the manifold-based matting methods generally need more E-time, because they need to construct patch alignment matrixes and use Nesterov's algorithm to optimize iteratively. Others are non-manifold based matting methods, and they need less E-time. Especially, Shared matting which has a real-time performance is implemented using C++.

Utilizing different manifold algorithms, the proposed manifold matting framework can obtain different foreground masks. The experiments above demonstrate that manifold matting algorithms have common inherent data traits for matting problem. So it is necessary and worthwhile to summary the existing manifold matting algorithms in an unified manifold matting framework. Both qualitative and quantitative comparisons can prove that the ISOMAP matting and CasISO matting method fit the matting framework and perform well. Additionally, some more matting results of the CasISO matting method on the dataset are provided in Figure. \ref{fig5}. Except for the 3 images showed in Figure. \ref{fig4}, we in this figure show the alpha results of other 24 images from the alphamatting dataset. For most of these images, the CasISO matting method can give excellent mask results. Therefore, all of the experimental results of CasISO matting method are showed in this paper. We can see the foreground matting results are smooth and continuous, such as the complex hairs of Barbie doll images and small inner holes of the flower and plant images. So the CasISO matting method is robust and also reveal the effectiveness of the proposed framework PAMM.

In summary, the above experiments demonstrate that the ISOMAP matting and CasISO matting methods resulted from our proposed manifold matting framework PAMM are effective and feasible compared with eight representative matting methods.

\section{Conclusions and Future Work}
\label{sec:5}

In this paper, we investigate the image matting problem, and propose a new patch alignment manifold matting (PAMM) framework and its two concrete algorithms, the ISOMAP matting  and its extension CasISO matting. This PAMM framework consists of part modeling and whole alignment optimization by minimizing the reconstruction error with the efficient Nesterov's algorithm. In addition, it is a unified manifold matting framework in which manifold learning methods can be incorporated as a manifold dimension reduction step. The experimental results show the effectiveness of the manifold matting framework. Moreover, the proposed example matting algorithms, ISOMAP matting and CasISO matting perform better than the several representative methods in some senses. In our future work, we will perform real-time manifold matting. On the other hand, we plan to use deep learning methods to perform image matting.

\bibliographystyle{IEEEtran}
\bibliography{IEEEfull,hao}

\begin{IEEEbiographynophoto}{Xuelong Li} (M'02-SM'07-F'12) is currently a Full Professor with the Center
for Optical Imagery Analysis and Learning, Xi'an Institute of Optics and Precision Mechanics,
Chinese Academy of Sciences, Xi'an, China.
\end{IEEEbiographynophoto}

\begin{IEEEbiography}[{\includegraphics[width=1in,height=1.25in,clip,]{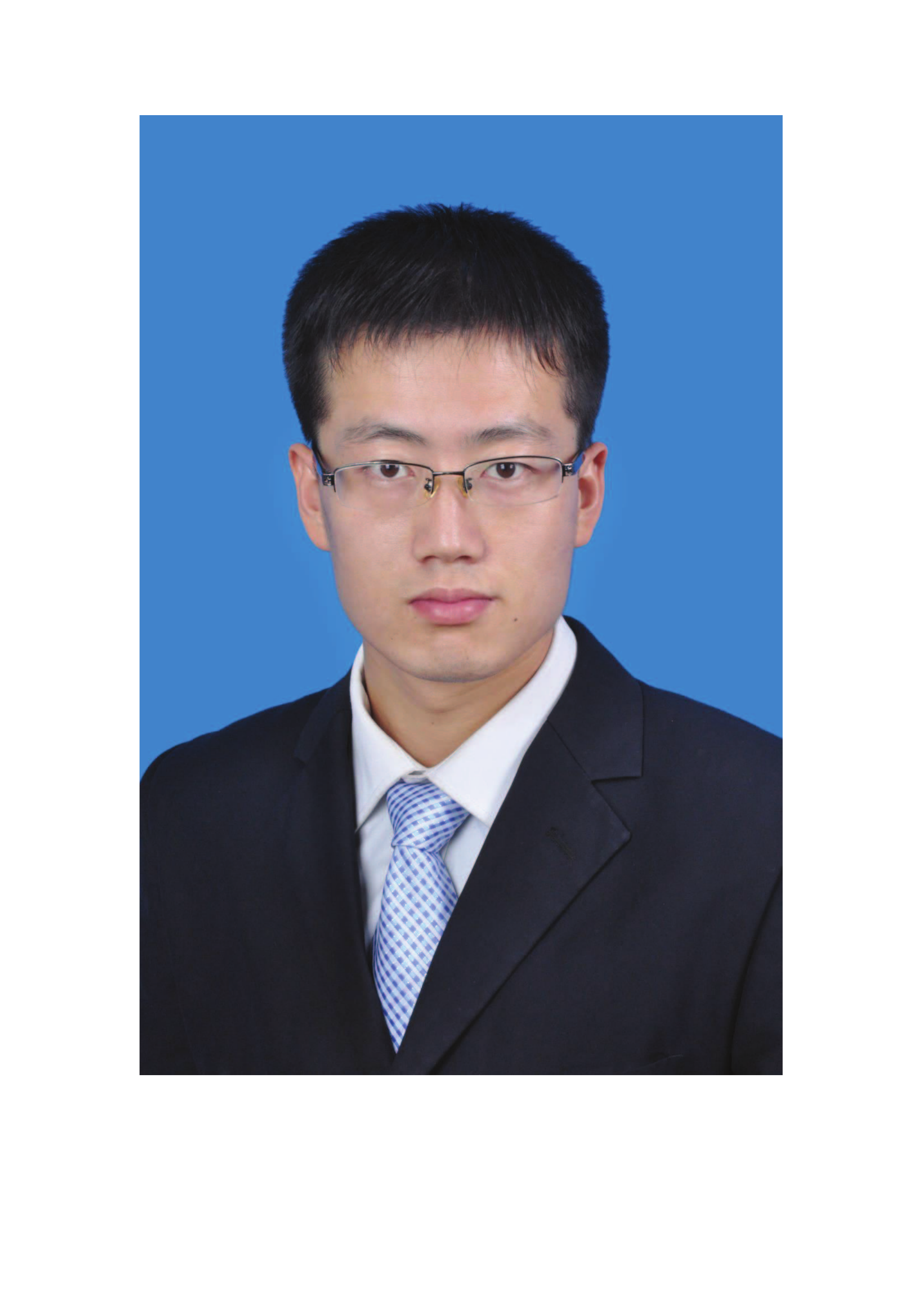}}]{Kang Liu} (M'17) received the Bachelor degree in computer science and technology from Xi'an Jiaotong University, Xi'an, China, in 2013, and received the Master degree in electronics and communication engineering from University of Chinese Academy of Sciences, Beijing, China, in 2016. He is currently an Assistant Research Fellow with the Center for OPTical IMagery Analysis and Learning (OPTIMAL), Xi'an Institute of Optics and Precision Mechanics, Chinese Academy of Sciences, Xi'an, China. His current research interests include pattern recognition, machine learning, and computer vision.
\end{IEEEbiography}

\begin{IEEEbiography}[{\includegraphics[width=1in,height=1.25in,clip,]{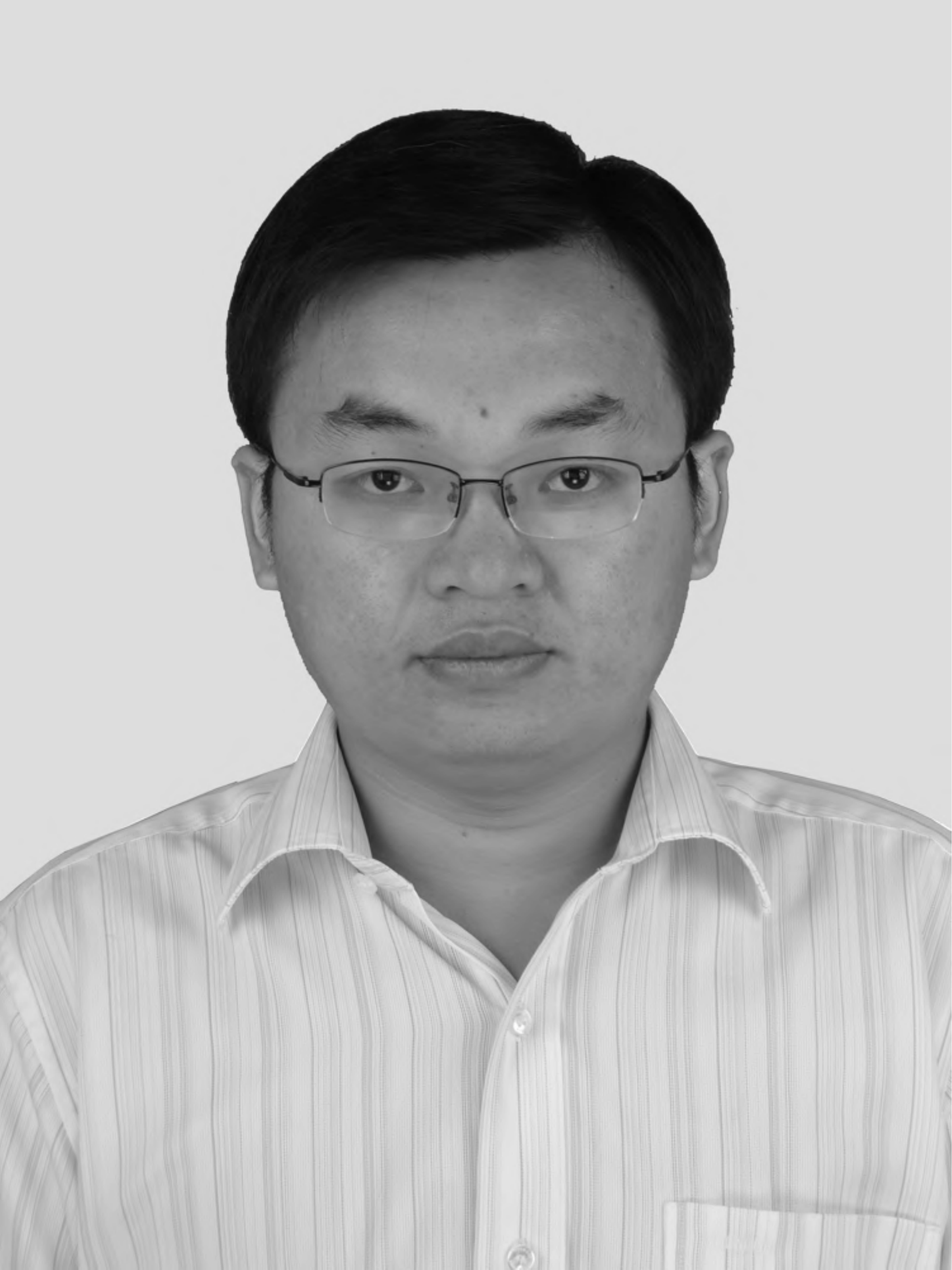}}]{Yongsheng Dong} (M'14) received the Ph.D. degree in applied mathematics from Peking University,
Beijing, China, in 2012. He was a Post-Doctoral Fellow with the Center for Optical Imagery Analysis and Learning, Xi'an Institute of Optics and Precision Mechanics, Chinese Academy of Sciences, Xi'an, China, from 2013 to 2016. He is currently an Associate Professor with the School of Information Engineering, Henan
University of Science and Technology, Luoyang, China. His current research interests include pattern recognition, machine
learning, and computer vision.  Dr. Dong is a member of the ACM and CCF. He has served as a Reviewer for
over 30 international prestigious journals and conferences, such as the IEEE T-NNLS, T-IP, T-CYB, and T-KDE.
He has also served as a Program Committee Member for over ten international conferences.
\end{IEEEbiography}

\begin{IEEEbiography}[{\includegraphics[width=1in,height=1.25in,clip,]{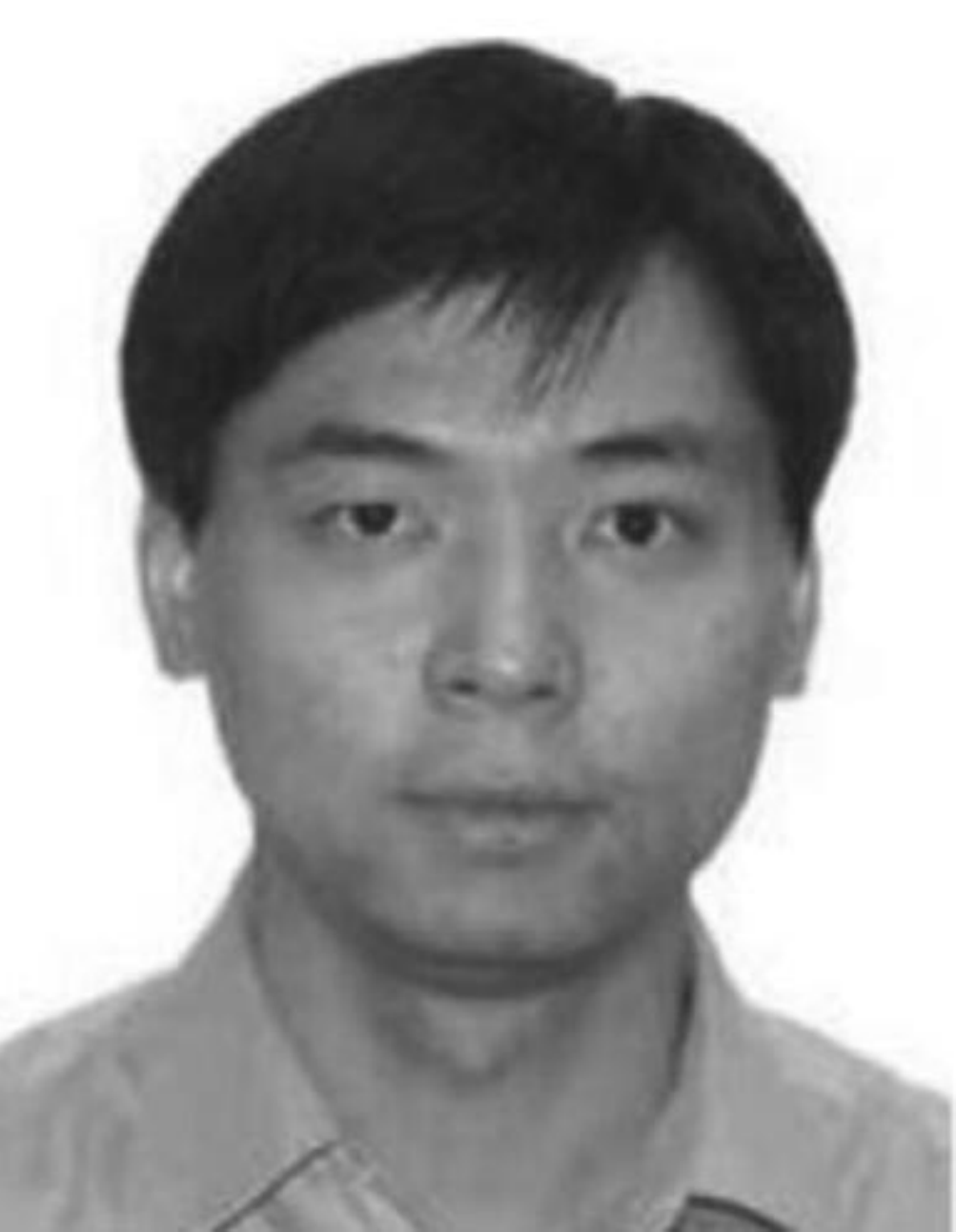}}]{Dacheng Tao} (F'15) is Professor of Computer Science and ARC Future Fellow in the School of Information Technologies and the Faculty of Engineering and Information Technologies, and the Inaugural Director of the UBTECH Sydney Artificial Intelligence Institute, at The University of Sydney. He mainly applies statistics and mathematics to Artificial Intelligence and Data Science. His research interests spread across computer vision, data science, image processing, machine learning, and video surveillance. His research results have expounded in one monograph and 500+ publications at prestigious journals and prominent conferences, such as IEEE T-PAMI, T-NNLS, T-IP, JMLR, IJCV, NIPS, CIKM, ICML, CVPR, ICCV, ECCV, AISTATS, ICDM; and ACM SIGKDD, with several best paper awards, such as the best theory/algorithm paper runner up award in IEEE ICDM'07, the best student paper award in IEEE ICDM'13, the 2014 ICDM 10-year highest-impact paper award, and the 2017 IEEE Signal Processing Society Best Paper Award. He received the 2015 Australian Scopus-Eureka Prize, the 2015 ACS Gold Disruptor Award and the 2015 UTS Vice-Chancellor's Medal for Exceptional Research. He is a Fellow of the IEEE, OSA, IAPR and SPIE.
\end{IEEEbiography}

\end{document}